\documentclass{article}

     \usepackage[nonatbib]{neurips_2020}

\usepackage[utf8]{inputenc} 
\usepackage[T1]{fontenc}    
\usepackage{hyperref}       
\usepackage{url}            
\usepackage{booktabs}       
\usepackage{amsfonts}       
\usepackage{nicefrac}       
\usepackage{microtype}      
  \usepackage{amsmath}
  \usepackage{epsfig,amssymb,subfigure,bm,dsfont}
  \usepackage{algpseudocode}
  \usepackage{amsfonts}
  \usepackage{array}
  \usepackage{balance}
  \usepackage{bbding}
  \usepackage{booktabs}
  \usepackage{calc}
  \usepackage{cite}
  \usepackage{color}
  \usepackage{curves}
  \usepackage{epstopdf}
  \usepackage{epic}
  \usepackage{float}
  \usepackage{footnote}
  \usepackage{graphicx}
  \usepackage{mdwlist}
  \usepackage{multicol}
  \usepackage{multirow}
  \usepackage{ntheorem}
  \usepackage{pifont}
  \usepackage[english]{babel}

\title{Blockchain Assisted Decentralized Federated Learning~(BLADE-FL) with Lazy Clients}

\author{%
        Jun~Li$^{\dag}$, Yumeng~Shao$^{\dag}$, Ming~Ding$^{\ddag}$, Chuan~Ma$^{\dag}$, Kang~Wei$^{\dag}$, Zhu Han$^{\S}$, H.~Vincent~Poor$^{\sharp}$ \\
        ${\dag}$ Nanjing University of Science and Technology, Nanjing, China\\
        $\ddag$ Data61, CSIRO, Sydney, Australia\\
        $\S$ University of Houston, Houston, TX, USA\\
        ${\sharp}$ Princeton University, NJ, USA\\
        \{jun.li, shaoyumeng, chuan.ma, kang.wei\}@njust.edu.cn\\
        ming.ding@data61.csiro.au \quad hanzhu22@gmail.com \quad poor@princeton.edu
}

\begin{document}

\newtheorem{definition}{Definition}
\newtheorem{assumption}{Assumption}
\newtheorem{lemma}{Lemma}
\newtheorem{theorem}{Theorem}
\newtheorem{remark}{Remark}
\newtheorem{proposition}{Proposition}

\maketitle

\begin{abstract}
Federated learning~(FL), as a distributed machine learning approach, has drawn a great amount of attention in recent years. FL shows an inherent advantage in privacy preservation, since users' raw data are processed locally. However, it relies on a centralized server to perform model aggregation. Therefore, FL is vulnerable to server malfunctions and external attacks.
In this paper, we propose a novel framework by integrating blockchain into FL, namely, blockchain assisted decentralized federated learning~(BLADE-FL), to enhance the security of FL. The proposed BLADE-FL has a good performance in terms of privacy preservation, tamper resistance, and effective cooperation of learning. However, it gives rise to a new problem of training deficiency, caused by lazy clients who plagiarize others' trained models and add artificial noises to conceal their cheating behaviors.
To be specific, we first develop a convergence bound of the loss function with the presence of lazy clients and prove that it is convex with respect to the total number of generated blocks $K$. Then, we solve the convex problem by optimizing $K$ to minimize the loss function. Furthermore, we discover the relationship between the optimal $K$, the number of lazy clients, and the power of artificial noises used by lazy clients.
We conduct extensive experiments to evaluate the performance of the proposed framework using the MNIST and Fashion-MNIST datasets. Our analytical results are shown to be consistent with the experimental results. In addition, the derived optimal $K$ achieves the minimum value of loss function, and in turn the optimal accuracy performance.
\end{abstract}

\section{Introduction}
\label{sec:introduction}

With the development of Internet of Things~(IoT) technology, the amount of data from user devices is exploding at an unprecedented rate~[1]. Traditional centralized technologies by collecting data from end devices are no longer suitable, due to the bottleneck of uploading bandwidth~[2]. To tackle this challenge, distributed machine learning~(DML) has emerged to achieve high reliability and low latency by processing distributive data at either end devices or the edge of networks~[3]. The DML can alleviate the burden on the central server by dividing a task into sub-tasks assigned to multiple nodes. However, DML needs to exchange samples when training a task~[4],
posing a serious risk of privacy leakage~[5]. To address this fundamental concern, federated learning (FL), proposed by Google~[6], begins to show its potential advantages. In a conventional FL system, a machine learning model is trained across multiple distributed clients with local datasets, and aggregated on a centralized server. By this approach,
FL is able to cooperatively complete machine learning tasks without directly sacrificing the data privacy~[7]. Currently, FL has been applied to many data-sensitive scenarios, such as smart health-care, e-commerce platform~[8], and the Google project Gboard~[9].

Although FL has shown its effectiveness on privacy protection~[10, 11], it still relies heavily on a single central server. Thus, FL is vulnerable to server malfunctions and external attacks, incurring inaccurate model updates, or even learning failures. In order to resolve this single-point-failure issue, blockchain~[12, 13] has been introduced to the FL systems~[14].
Taking the advantage of blockchain, the work in~[14] developed a blockchain-enabled FL architecture to validate the uploaded parameters and investigated the related system performances, such as block generating rate and learning latency. The work in~[15] applied \textit{Delegated Proof of Stake}~(DPoS) into blockchained FL to reduce the delay at the expense of robustness. The work in~[16] developed a tamper-proof architecture that uses blockchain to enhance the security when sharing parameters, and proposed a new consensus mechanism, namely, \textit{Proof of Quality}~(PoQ), to determine the rewards allocation in the blockchain.
Although these works replaced the central sever with blockchain to avoid single-point-failure, they inevitably introduced third-party devices, known as miners in blockchain, to store the aggregated models in a distributed manner.
This will cause the model leakage issue, since the model parameters are open to the miners in the blockchain.
In addition, these works did not theoretically analyze the convergence performance of loss function, and optimize the computational resources allocation, in Blockchained FL.

In this work, we propose a novel framework, namely, blockchain assisted decentralized FL~(BLADE-FL). In this framework, both training and mining processes are executed at the client-side, i.e., each client will conduct both the model training and block generating tasks. We consider a synchronous case that all the clients follow the same time allocation scheme for local training and block generating. We pay special attention to the phenomenon that part of the clients, known as \emph{lazy clients}, may tend to save their computing resources by directly copying models from others, leading to training deficiency and performance degradation.

Based on the above discussion, the main contributions of this paper can be summarized as follows.

\begin{itemize}
        \item {We propose a novel blockchained FL framework, known as BLADE-FL, which has a good performance on privacy preservation, and effective cooperation of learning, compared with conventional blockchained FL.
            }
        \item {We perform theoretical analysis on the upper bound of the loss function with the presence of lazy clients, to evaluate the learning performance of BLADE-FL, and prove that it is convex with respect to the total number of generated blocks $K$.
            }
        \item{We solve the convex problem by optimizing $K$ to minimize the loss function, and further discover the relationship between the optimal $K$, the number of lazy clients, and the power of artificial noises.
        }
        \item{ We validate our theoretical results with the experimental results. It is shown that the derived optimal $K$ captures the minimum value of loss function and achieves the optimal accuracy performance. }
    \end{itemize}

\section{Background}\label{sec:Rela_Work}
\subsection{Federated Learning}

As a distributed machine learning, FL is designed to develop efficient machine learning between multiple participants under the premise of information security, protecting the privacy of personal data. To analyze the performance of FL, the work in~[17] derives an upper bound of the loss function between the iterations of local training and global aggregation.

Assume there are $N$ clients, with \emph{non-IID} dataset $\mathcal D_i$ at $i$-th client, $i=1,2,\dots,N$, and each client has the same number of local dataset.
Therefore, we define the global loss function as:
$F(\boldsymbol{w})\triangleq \frac{1}{N}\sum_{i=1}^N F_i(\boldsymbol{w})$,
where $F_i(\cdot)$ is the local loss function.
In FL, each client is trained locally to minimize the local loss function, while the entire system is trained to minimize the global loss function $F(\boldsymbol{w})$.
Global aggregation is performed through an aggregator to average the weights of each node, then updates the global model:
$\boldsymbol{w}^{k}=\frac{1}{N}\sum_{i=1}^N \boldsymbol{w}_{i}^{k}$,
where $\boldsymbol{w}_{i}^{k}$ and $\boldsymbol{w}^{k}$ denote the local parameters in $i$-th client and the aggregated parameters, respectively, at $k$-th communication round.
The learning process will end until $\triangle F(\boldsymbol{w})$=$F(\boldsymbol{w}^k)-F(\boldsymbol{w}^{k-1})<\epsilon$, and $\epsilon$ is an arbitrary small number.

\subsection{Blockchain}
Blockchain is a shared and decentralized ledger. The data stored in the blockchain is considered immutable, thanks to the consensus mechanism. The consensus mechanism validates the data within the blocks and ensures that all the nodes participating in the blockchain store the same data. Thus, the data stored in the blockchain is able to be accessed by any participating nodes and will remain invariable. The most prevalent consensus mechanism is \emph{Proof of Work}~(PoW). PoW is a mathematical problem that is easy to verify but extremely hard to solve, namely, solving these problems will consume a lot of computational resources. In addition, PoW requires the nodes in blockchain to play the role as a miner, doing the process called mining, to verify the data within the block. The mining process is a competition, that is finding a nonce which makes the hash value of the data meet the specific requirement. The hash value is calculated by the hash function, and different blockchain has different hash functions. For instance, the bitcoin system uses SHA256 hash function~[13], while Ethernet uses Keccak256~[19]. Due to the mining process, PoW can defense attacks on the condition that the total computational resources of malicious devices are less than the sum of honest devices~(51\% attack)~[20].

The blockchain constructs information in a chain of blocks, where each block stores a group of transactions. The blocks are linked together to form a chain by referencing the hash value of the previous block. Any change of the data within the blocks will result in a completely different hash value, destroying the preceding chain structure. Therefore, it is impossible to tamper the data that already stores in the blockchain.
Based on the above discussion, blockchain is safe and reliable with the consensus mechanism and chain structure, and thus can defend against external attacks.

\section{The Proposed Framework}\label{sec:System model}
In this section, we show our proposed BLADE-FL framework, the model of lazy clients, and the computational resources allocation model in detail.

\begin{figure}[ht]
  \centering
  \includegraphics[width=1\textwidth,height=0.86\textwidth]{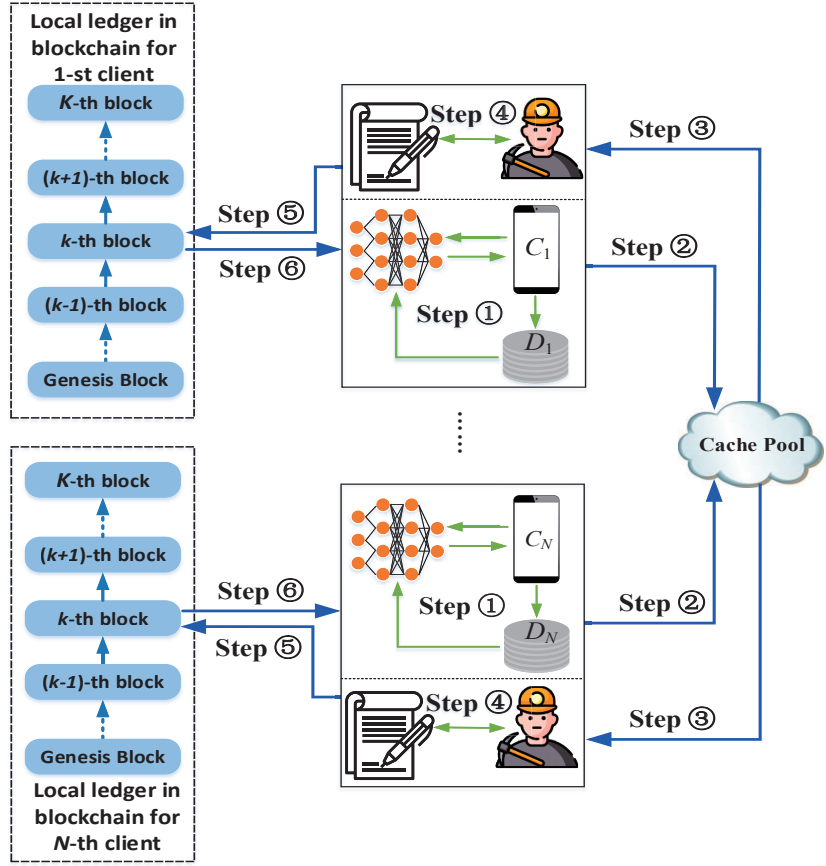}\\
  \caption{The procedure of the proposed BLADE-FL in each communication round, where the model training function and the mining function are collocated at each client. The process is for the $k$-th block, corresponding to $k$-th communication round in FL. The total number of generated blocks $K$ is equal to that of the communication rounds in FL, numerically.
  }\label{graph_framework}
\end{figure}

\subsection{BLADE-FL}
We consider a synchronous scenario, where all the clients begin to train local models at the same time, and then turn to generate blocks simultaneously.
Firstly, all the clients initialize their parameters, such as model initial weight, learning rate, etc.
After initialization, all the clients will go through the procedure of training and mining between two generated blocks. We use $K$ to represent the total number of generated blocks, which is also equal to the number of communication rounds.
We divide the procedure into the following steps in each communication round, as illustrated in Fig.~\ref{graph_framework}:

\begin{basedescript}{\desclabelstyle{\pushlabel}\desclabelwidth{4em}}
\item[$\bullet$ \textbf{Step 1}:]
\emph{Local Training.} Each client trains local samples to update its own model $\boldsymbol{w}_i^k, \forall i$.
\item[$\bullet$ \textbf{Step 2}:]
\emph{Model Uploading.} All the clients encrypt their models by homomorphic encryption and uploads the models into the cache pool. The cache pool is a database that only stores the models provisionally.
\item[$\bullet$ \textbf{Step 3}:]
\emph{Model Downloading.} Clients download all the models in the cache pool, but do not aggregate them directly. This is because all the models in the cache pool are not verified, and it is possible to be tampered by external malicious clients. Only after the mining, a new block that included all the verified models has been generated in the local ledger of each client, causing the models immutable and reliable.
\item[$\bullet$ \textbf{Step 4}:]
\emph{Mining.} Clients start mining by adding a random number at the end of downloaded models and calculate the hash value. Miners will repeat this produce until the nonce~(the random number that the calculated hash value satisfies the block generating difficulty) has been found or another client has broadcast the nonce.
\item[$\bullet$ \textbf{Step 5}:]
\emph{Block Verification.} Once a client has found the nonce, it broadcasts the nonce to publish a new block. Other clients will verify the nonce by their own. When more than half of clients confirm the number, a new block has been published. Then all clients in the system will append the new block onto their local ledgers.
\item[$\bullet$ \textbf{Step 6}:]
\emph{Local Updating.} We notice that the new block contains all the models that ready to be aggregated. Thus, the global aggregation is processed locally when a new block is generated. Thereafter, the $k$-th generated block is equal to the $k$-th communication round, and the total number of generated blocks $K$ denotes $K$ communication rounds, numerically. That is, all the clients download the models in the new block to generate the global model, and update their local models with the aggregated global model $\boldsymbol{w}^k$ for the $k$-th block. Afterward, the system finishes the procedure in the $k$-th block, and turn to the next block until the final $K$-th block~(computational resources is run off).
\end{basedescript}

The BLADE-FL has a good performance on privacy preservation, since there is no need to introduce an additional third-party for block generating, reducing the possibility of model leakage.
Besides, the BLADE-FL is tamper-resisted, thanks to the consensus mechanism.

\subsection{Model of Lazy Clients}
Different from traditional FL, a new problem of learning deficiency caused by lazy clients emerges in the BLADE-FL system. This issue is fundamentally originated from the lack of an effective detection and penalization mechanism in an unsupervised network. In more detail, a lazy client can plagiarize other models directly before generating a new block. To avoid being spotted by the system, they will add artificial noises to the model weights.
Such a cheating behavior can be expressed as:
\begin{equation}
\boldsymbol{w}_{i'}^{k}=\boldsymbol{w}_i^{k}+\mathbf{n}_i,\quad i'\subseteq \mathcal{M}, i\nsubseteq \mathcal{M}, k=1,2,\dots,K,
\end{equation}
where $\mathcal{M}$ denotes the set of lazy clients, $\mathbf{n}_i$ is the artificial noise vector following a Gaussian distribution with a zero mean and a variance of $\sigma^2$.

\subsection{Computational Resources Allocation Model}

The generating rate of blockchain is determined by the computation complexity of the hash function and available computational resources.
The computational resources of CPU cycles required to generate a block in PoW is defined as: $\mathbb{E}[\mathrm{PoW}]=\kappa\chi$, where $\kappa$ is the block generating difficulty to stabilize the block generating rate, and $\chi$ denotes the amount of required computational resources for calculating one hash value~[21]. Thus, we define the time for mining a block in terms of $\beta$:
\begin{equation}\label{eq_Etbl}
\begin{aligned}
\beta \triangleq \mathbb{E}[t^{\mathrm{bl}}]=\frac{\mathbb{E}[\mathrm{PoW}]}{f^{\mathrm{bl}}}
=\frac{\kappa\chi}{f^{\mathrm{bl}}},
\end{aligned}
\end{equation}
where $f^{\mathrm{bl}}$ denotes the CPU cycles of the whole blockchain network for block generating.

Similarly, we can use the computational resources of CPU cycles required for every single sample training $\rho$, to derive the time that needed to train all the local samples for one round at $i$-th client, in terms of $\alpha_i$~[22]:
\begin{equation}\label{eq_Etloc}
\alpha_i \triangleq \mathbb{E}[t^{\mathrm{loc}}]=\frac{\vert \mathcal{D}_i\vert \rho}{f^{\mathrm{loc}}},
\end{equation}
where $f^{\mathrm{loc}}$ denotes the CPU cycles for local training and $\vert \mathcal{D}_i\vert$ denotes the number of local dataset.
In order to keep the same structure of the models, we assume that each client uses the same algorithm to train the local model and has the same number of samples.
Thereafter, the local training time per round is the same for all the clients, where we define $\alpha=\alpha_i, \forall i$.

In fact, the computational resources of each client are limited in the total computing time $t^{\mathrm{sum}}$, and local training in FL and block generating in blockchain both require computational resources. Thus, it is crucial to allocate computing resources between local training and block generating.
Based on the above discussion of the time for training per round $\alpha$ and the time for mining per block $\beta$, the computational resource allocation can be transformed into the time allocation:
\begin{equation}\label{eq_t_allocation}
K\tau\alpha+K\beta=t^{\mathrm{sum}},
\end{equation}
where $\tau$ represents the number of training iterations between two generated blocks, and $K\tau\alpha$ is the training time while $K\beta$ is the block generating time.

\section{Performance Analysis of the BLADE-FL with Lazy Clients}\label{sec:Con_FL}

We assume there are $N$ clients in the BLADE-FL system, while each client needs to train local samples in FL and acts as a miner to generate blocks in blockchain.
\subsection{Performance Analysis on Upper Bound}
To analyze the performance of the proposed framework, we make the following assumptions of the loss function.
\begin{assumption}\label{assumption_1}
\textit{We assume the followings for all the clients:}\\
\textit{1) $F_i(\boldsymbol{w})$ is convex that}
$F(\boldsymbol{w})\geq F(\boldsymbol{w}')+\nabla F(\boldsymbol{w}')^T (\boldsymbol{w}-\boldsymbol{w}')$;\\
\textit{2) $F_i(\boldsymbol{w})$ is $\xi$-Lipschitz that}
$\Vert F_i(\boldsymbol{w})-F_i(\boldsymbol{w}')\Vert\leq \xi \Vert\boldsymbol{w}-\boldsymbol{w}'\Vert$, for any $\boldsymbol{w},\boldsymbol{w}'$;\\
\textit{3) $F_i(\boldsymbol{w})$ is L-smooth that}
$\Vert\nabla F_i(\boldsymbol{w})-\nabla F_i(\boldsymbol{w}')\Vert\leq L \Vert\boldsymbol{w}-\boldsymbol{w}'\Vert$, for any $\boldsymbol{w},\boldsymbol{w}'$.
\end{assumption}

According to \textbf{Assumption~\ref{assumption_1}}, $F(\boldsymbol{w})$ is \textit{convex, $\xi$-Lipschitz,} and \textit{L-smooth}~[23].

In order to capture the divergence between the global loss function and the gradient of the local loss function, we also define the following measurements.
\begin{definition}
(Gradient Divergence) For any client, we define $\delta_i$ as an upper bound, $\Vert\nabla F_i(\boldsymbol{w})-\nabla F(\boldsymbol{w})\Vert\leq \delta_i$. Thus, the global gradient divergence $\delta$ can be expressed as $\delta=\frac{\sum_i \vert\mathcal D_i\vert \delta_i}{N}$.
\end{definition}

Then, we develop the upper bound of the loss function in the BLADE-FL without lazy clients, regarded to the total number of generated blocks $K$ as the following Lemma shows.
\begin{lemma}
The convergence upper bound of BLADE-FL after $K$ blocks without lazy clients is given by:
\begin{equation}\label{lemma_equation}
F(\boldsymbol{w}^{K})-F(\boldsymbol{w}^*)\leq
\frac{1}{\gamma \left(\eta\phi-\frac{\frac{\delta\xi K}{L}\left( \lambda^{\frac{\gamma}{K}}-1\right)-\eta\xi\delta \gamma}{\varepsilon^2 \gamma}\right)},
\end{equation}
where $\varepsilon>0$, $\lambda=\eta L+1$, $\gamma=\frac{t^{\mathrm{sum}}-K\beta}{\alpha}$, and $\phi=\frac{(1-\frac{\eta L}{2})}{\Vert\boldsymbol{w}(0)-\boldsymbol{w}^*\Vert}$, $\boldsymbol{w}^0$ denotes the initial weight, $\boldsymbol{w}^*$ denotes the optimal weight, $\eta$ denotes the learning rate and $\eta L<1$.
\end{lemma}

Based on \textbf{Lemma~1}, we develop the upper bound of BLADE-FL with lazy clients in the following theorem.

\begin{theorem}
The upper bound of BLADE-FL with $M$ lazy clients is given by:
\begin{equation}\label{final_equation}
F(\boldsymbol{w}^{K})-F(\boldsymbol{w}^*) \leq \frac{1}{\gamma \left(\eta\phi-\frac{\frac{\delta\xi K}{L}\left( \lambda^{\frac{\gamma}{K}}-1\right)-\eta\xi\delta \gamma+\xi\frac{M}{N}\theta+\xi\frac{\sqrt{M}}{N}\sigma^2}{\varepsilon^2 \gamma}\right)},
\end{equation}
where $\theta$ denotes the degradation of the system performance caused by lazy clients and can be represented by: $\theta = \Vert\boldsymbol{w}_{i'}^{\mathrm{nolazy}}-\boldsymbol{w}_{i'}^{\mathrm{lazy}}\Vert$.
\end{theorem}

Thereafter, we use the upper bound in~(\ref{final_equation}) to represent the learning performance of the BLADE-FL with lazy clients.

\subsection{Discussion on Performance}\label{Sec:Lazy}

The developed upper bound indicates that the learning performance depends on the total number of generated blocks $K$, the time for training local samples per round $\alpha$, the time for mining per block $\beta$, and the total computing time $t^{\mathrm{sum}}$.
In this paper, we assume that $\alpha$ and $\beta$ are constant since they are usually  configured by the BLADE-FL system to ensure a synchronous protocol, and we only concentrate on the total number of generated blocks $K$.
Hence, the upper bound only relies on the total number of generated blocks $K$, since $t^{\mathrm{sum}}$ is usually cast as a resource constraint. Then we obtain the following remark.
\begin{remark}
The developed upper bound is a convex function with respect to $K$, so we can solve the convex function to find the optimal $K$ to minimize the loss function of the BLADE-FL with lazy clients.
\end{remark}

In the developed upper bound shown in (\ref{final_equation}), we notice that $\sigma^2$ should have a similar order of magnitude compared with $\theta$, because a lazy client tends not to add a huge or tiny noise in order to conceal itself.
Therefore we can obtain the following remark.
\begin{remark}
In the lazy BLADE-FL system, the plagiarism behavior contributes a term proportional to $\frac{M}{N}$ to the bound while the artificial noise exhibits an impact term proportional to $\frac{\sqrt{M}}{N}$. This is because the plagiarism behavior is related to the original data while the artificial noise is independent of the original data, which indicates the plagiarism has a greater effect on the learning performance compared with the noise perturbation.
\end{remark}

Then, we reveal the impact of $\frac{M}{N}$ and $\sigma^2$ on the optimal value of $K$ in the following remark.

\begin{remark}
The optimal $K$ that minimizes the loss function value decreases as the lazy client ratio $\frac{M}{N}$ or the noise variance $\sigma^2$ grows. Intuitively speaking, when the system is infested with a large number of lazy clients, more computational resources should be redirected and allocated to FL to compensate for the insufficient computations of learning, and thus improving the learning performance.
\end{remark}

\section{Simulation Results}\label{sec:Exm_Res}
In this section, we evaluate the analytical results. First, we evaluate the gap between our developed upper bound and the simulation results. Then, we demonstrate the optimal $K$ of various lazy client ratio $\frac{M}{N}$ and various noise variance $\sigma^2$ in a limited computing time $t^{\mathrm{sum}}$.

\subsection{Experimental setting}

1) Datasets: In our experiments, we use two datasets for \textit{non-IID} setting to verify the results.

$\bullet$ MNIST. Standard MNIST handwritten digit recognition dataset consists of 60,000 training examples and 10,000 testing examples~[24]. Each example is a 28$\times$28 sized handwritten digits grayscale image from 0 to 9.

$\bullet$ Fashion-MNIST. Fashion-MNIST for clothes has 10 different types, such as T-shirt, trousers, pullover, dress, coat, sandal, shirt, sneaker, bag, and ankle boot.

2) MLP. We progress the learning of a Multi-Layer Perceptron~(MLP) model. The MLP network with a single hidden layer contains 256 hidden units. Each unit applies softmax function and rectified linear units of 10 classes~(corresponding to the 10 digits and 10 clothes). We use the cross-entropy to model the loss function, which can capture the error of the training data.

3) PoW. We evaluate the block generating process of PoW consensus mechanisms with the computational resources consumption. We assume that generating a block requires $\beta$ units time.

4) Parameters setting. In our experiments, we set the total computing time $t^{\mathrm{sum}}=100$, the samples of each client $\vert\mathcal{D}_i\vert=512,\forall i$, the number of clients $N=20$, time unit for local training per round $\alpha=1$, time units for mining per block $\beta=10$, number of lazy clients $M=0$ and learning rate $\eta=0.01$ as default.

\subsection{Evaluating the Performance with Lazy Clients}
In this subsection, we progress the learning task with lazy clients using MNIST and Fashion-MNIST datasets.
Then we search the optimal $K$ to obtain the best learning performance on various lazy client ratios $\frac{M}{N}$ and various noise variance $\sigma^2$.

\begin{figure}[t]
  \centering
  \subfigure[MNIST]{
  \begin{minipage}[t]{0.48\textwidth}
  \centering
  \includegraphics[height=0.8\textwidth,width=1\textwidth]{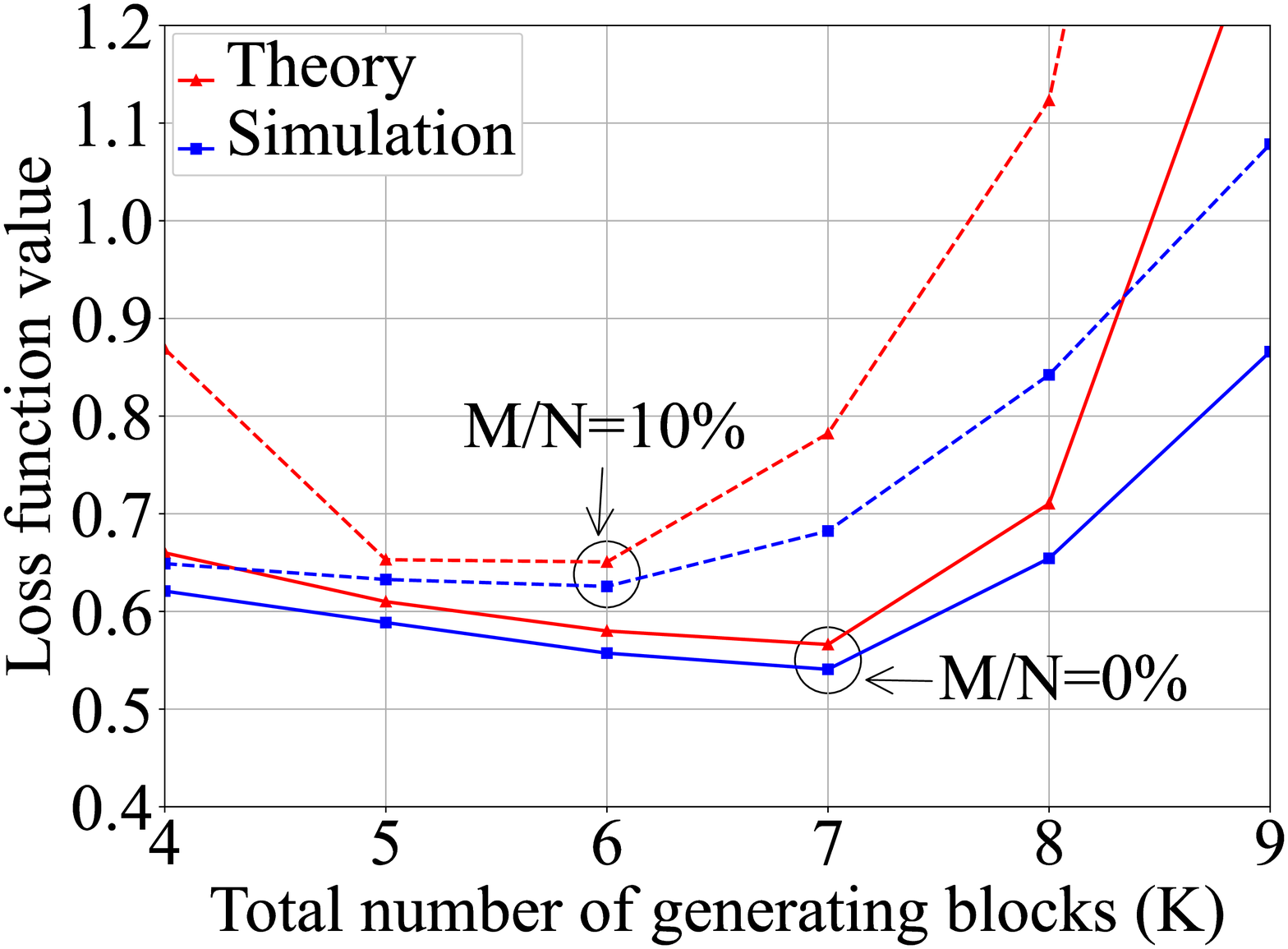}
  \end{minipage}
  }
  \subfigure[FashionMNIST]{
  \begin{minipage}[t]{0.48\textwidth}
  \centering
  \includegraphics[height=0.8\textwidth,width=1\textwidth]{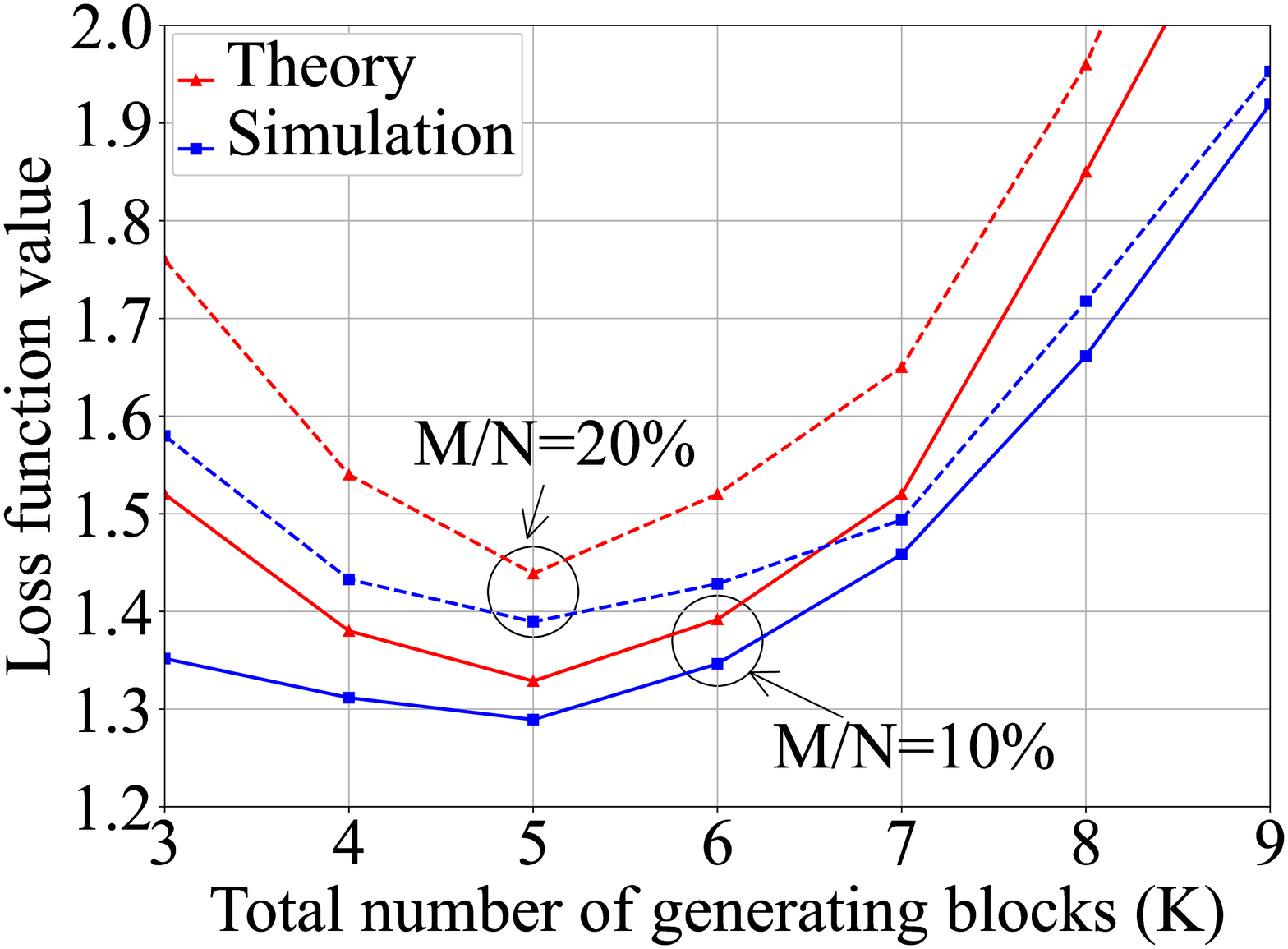}
  \end{minipage}
  }
  \caption{Numerical results and experimental results on MNIST and Fashion-MNIST}
    \label{fig_3}
\end{figure}

1) \textbf{Numerical Results.} We draw the numerical results in~(\ref{final_equation}) compared with the simulation results in Fig.~\ref{fig_3}. Fig.~\ref{fig_3} shows that our developed upper bound is consistent with the simulation result and can be a criterion to evaluate the learning performance. Thereafter, finding the optimal $K$ in the developed upper bound will achieve the best learning performance in the BLADE-FL.

\begin{figure}[t]
  \centering
  \subfigure[Loss function]{
  \begin{minipage}[t]{0.48\textwidth}
  \centering
  \includegraphics[height=0.8\textwidth,width=1\textwidth]{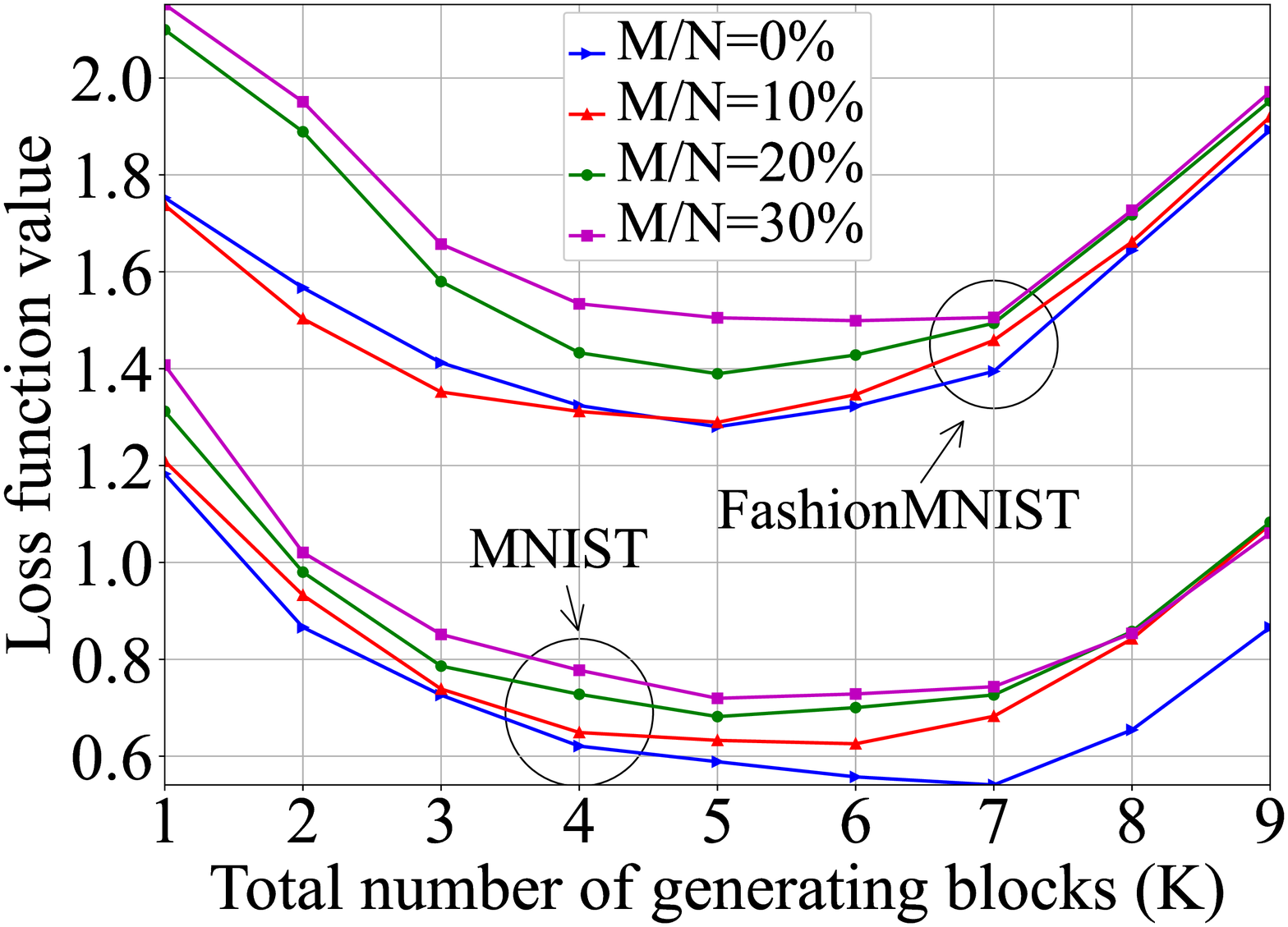}
  \end{minipage}
  }
  \subfigure[Accuracy]{
  \begin{minipage}[t]{0.48\textwidth}
  \centering
  \includegraphics[height=0.8\textwidth,width=1\textwidth]{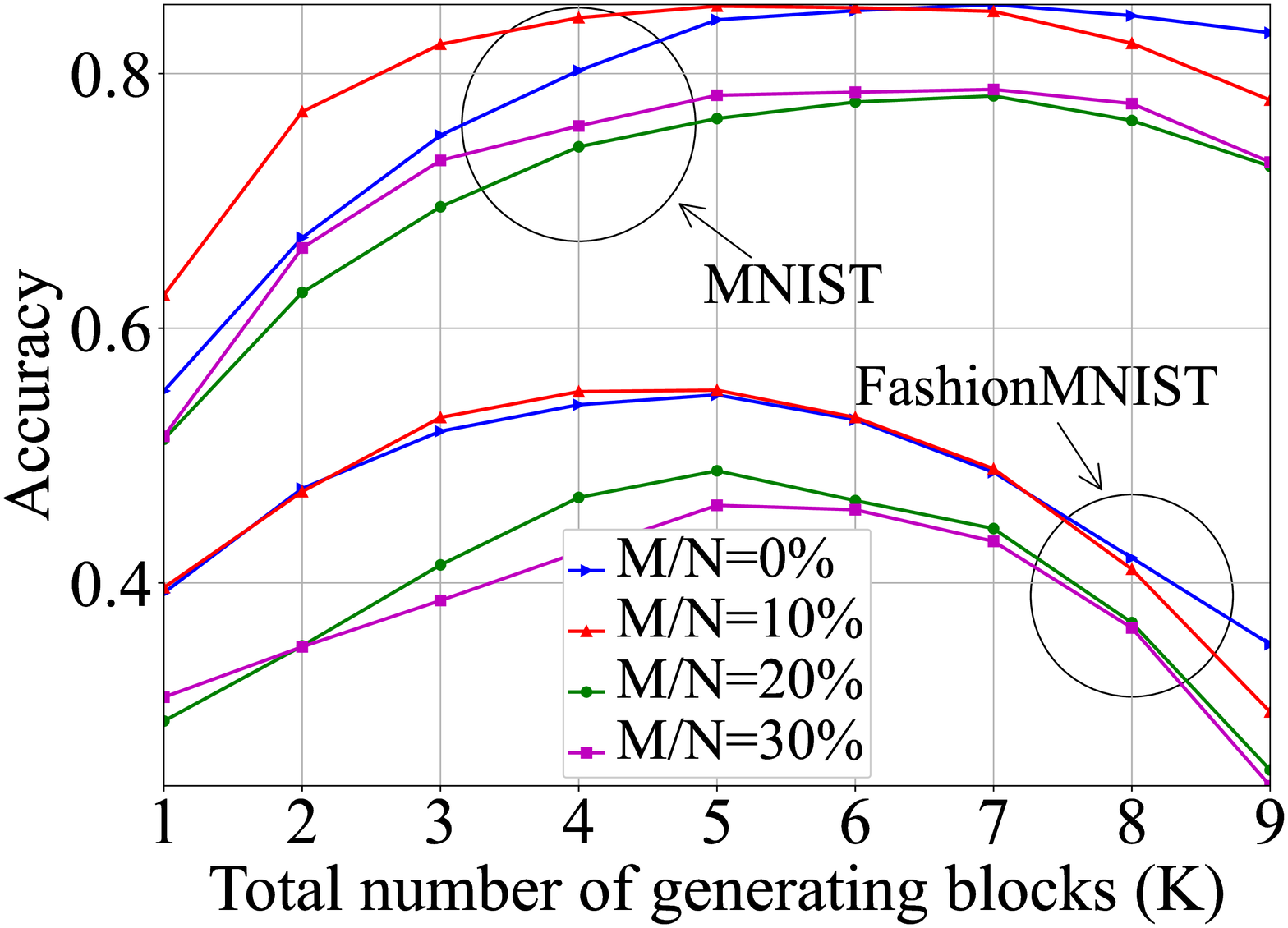}
  \end{minipage}
  }
  \caption{Value of the loss function and accuracy under various $K$ and $\frac{M}{N}$ values for the MNIST and FashionMNIST datasets}
    \label{fig_8}
\end{figure}

2) \textbf{Various Lazy Client Ratio.} We simulate various lazy clients~($M=0, M=2, M=4$, and $M=6$) when the total clients $N=20$, which makes the lazy client ratio $\frac{M}{N}$ to be $0\%, 10\%, 20\%$, and $30\%$. Other parameters are set as default, and we set the noise variance $\sigma^2=0.01$. In Fig.~\ref{fig_8}, we notice that a ascending lazy client ratio will lead to a descending optimal $K$~(from MNIST: $M=0,K^*=7;M=2,K^*=6;M=6,K^*=5$), which causes more computational resources allocated in local training.
Moreover, the learning performance will drop when more lazy clients are involved in the system.
Besides, the optimal $K$ on Fashion-MNIST also follows the changes in learning performance on MNIST. Therefore, our performance analysis with lazy clients can be applied to various learning datasets.

\begin{figure}[t]
  \centering
  \subfigure[Loss function]{
  \begin{minipage}[t]{0.48\textwidth}
  \centering
  \includegraphics[height=0.8\textwidth,width=1\textwidth]{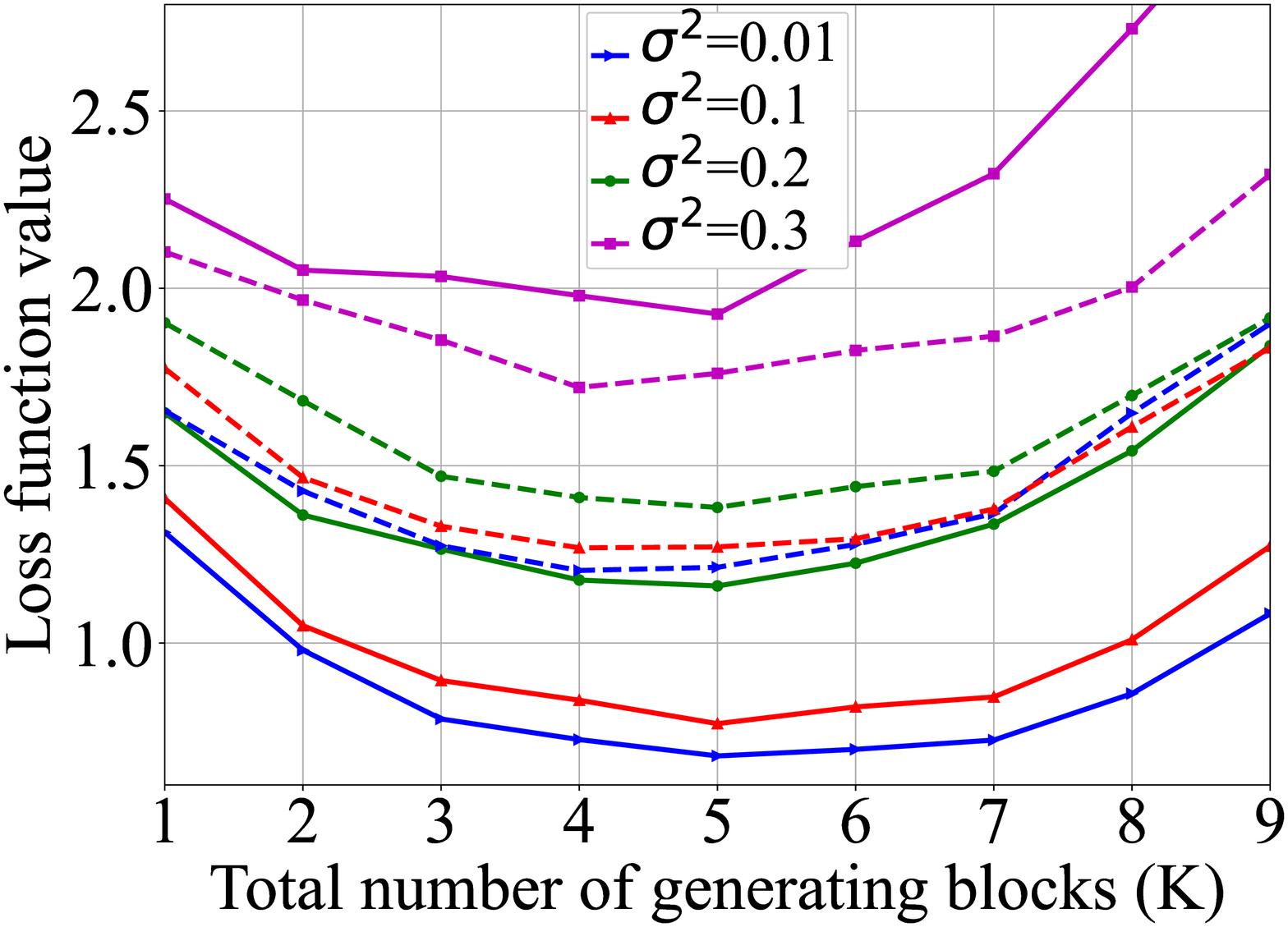}
  \end{minipage}
  }
  \subfigure[Accuracy]{
  \begin{minipage}[t]{0.48\textwidth}
  \centering
  \includegraphics[height=0.8\textwidth,width=1\textwidth]{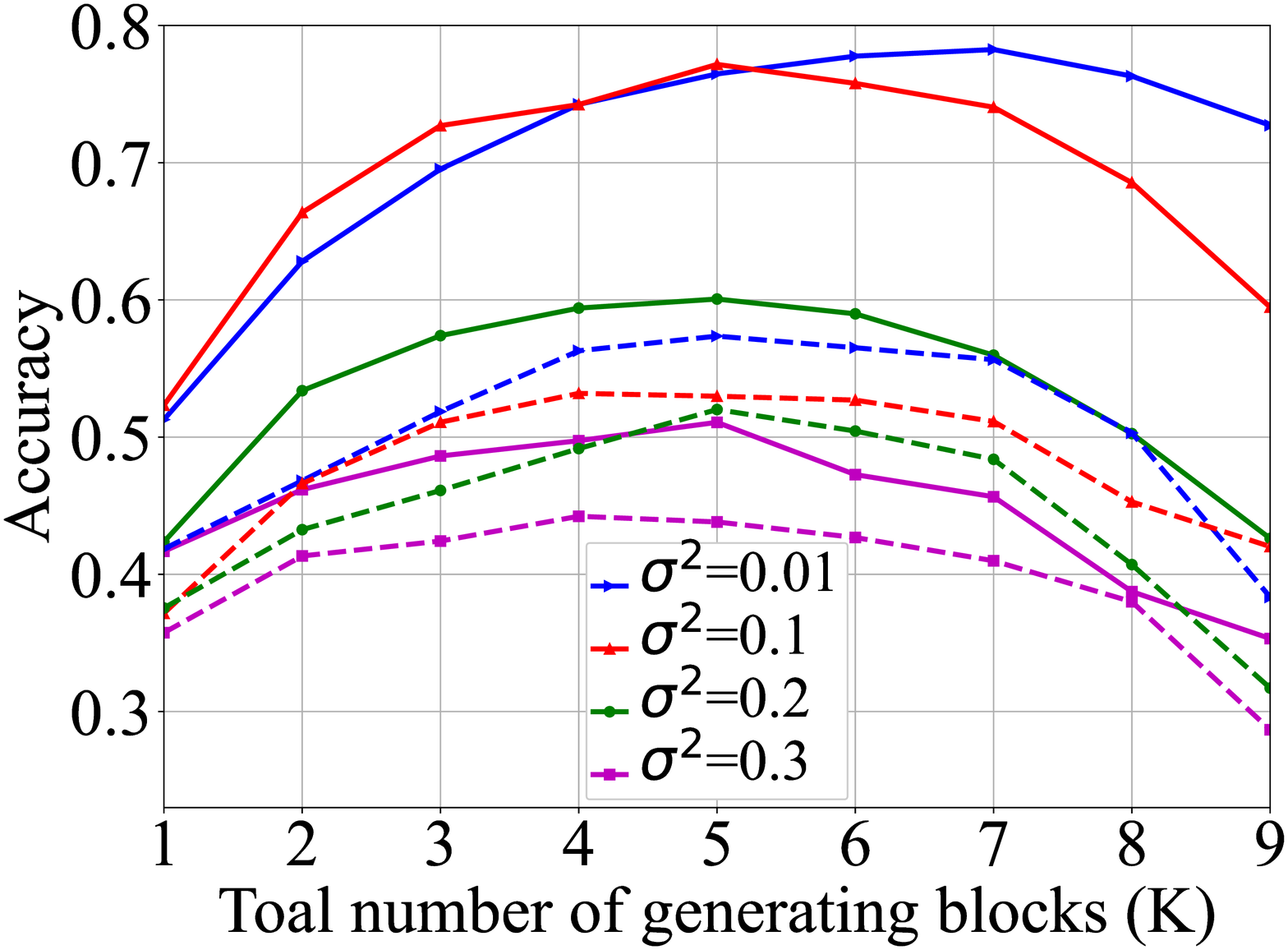}
  \end{minipage}
  }
  \caption{Value of the loss function and accuracy under various $K$ and $\sigma^2$ values on 20\% lazy ratio. Solid line is the results on MNIST while dash line on Fashion-MNIST}
  \label{fig_9}
\end{figure}

3) \textbf{Various Noise Variance.} In Fig.~\ref{fig_9}, we set different values of noise variance~($\sigma^2=0.01$, $\sigma^2=0.1$, $\sigma^2=0.2$, and $\sigma^2=0.3$) on lazy client ratio $\frac{M}{N}=20\%$ and plot their experiment results.
We notice that when noise variance $\sigma^2$ is large, the BLADE-FL can be damaged severely, which results in bad learning performance.
Furthermore, the optimal $K$ also descends as the noise variance increasing~(from MNIST, full line in Fig.~\ref{fig_9}: $\sigma^2=0.01,K^*=7; \sigma^2=0.3,K^*=5$), which means the BLADE-FL will allocate more computational resource in training with a high $\sigma^2$.

\section{Conclusion}\label{sec:Concl}
In this paper, we have proposed a BLADE-FL framework, with a good performance in terms of privacy preservation, tamper-resisted, and efficient cooperation of learning.
In order to investigate a unique problem in the proposed BLADE-FL system, called the Lazy Client Problem, we have theoretically developed a convergence bound of the loss function in the BLADE-FL system with the presence of lazy clients. We have proved the bound to be convex with respect to the total number of generated blocks $K$, and solved the convex problem to minimize the loss function by optimizing $K$.
Our experiments have demonstrated that the optimal $K$ varies with the number of lazy clients and the power of artificial noises, which is consistent with our theoretical results.

However, the lazy client problem is not been completely solved in this paper. In our future work, we are going to further investigate the impact of lazy clients and artificial noise. Then we will design effective algorithms to detect the plagiarism behaviors, and control the ratio of lazy clients.

\section*{Acknowledgments}
This work is supported in part by National Key R\&D Program under Grants 2018YFB1004800, in part by National Natural Science Foundation of China under Grants 61727802, 61872184.

\section*{Broader Impact}
Researchers who are interested in federated learning, blockchain, distributed learning, and AI will benefit from this research. Since our experimental results were obtained using non-sensitive datasets of handwritten digits and clothing, there is no ethical issue by disclosing our research outcome to a general audience. The investigated problem in a blockchain integrated FL system, if not treated properly, will degrade the learning performance, and lead to an unreliable system. The bias of the data is not applicable to our work since we investigated a general methodology in this paper, which does not rely on any particular distribution of data.

\section*{References}
\small

[1] Z. Yong-qun. \ (2018) Research on system integration based on big
data,internet of things and cloud computing technology. {\it Telecom
Power Technology}.

[2] R. Rajiv, R. Omer, N. Surya, Y. Mazin, J. Philip, Z. Wen, B. Stuart,
W. Paul, J. P. Prakash, and G. Dimitrios.\ (2018) \ The next grand challenges:
Integrating the internet of things and data science. {\it IEEE
Cloud Computing}, vol. 5, no. 3, pp. 12–26.

[3] X. Li, K. Li, D. Qiao, Y. Ding, and D. Wei.\ Application research of
machine learning method based on distributed cluster in information
retrieval. \ (2019) In {\it International Conference on Communications,
Information System and Computer Engineering (CISCE)}, pp. 411–414.

[4] H. B. Mcmahan, E. Moore, D. Ramage, S. Hampson, and B. A. Y.
Arcas.\ (2016) Communication-efficient learning of deep networks from
decentralized data.

[5] S. Shaham, M. Ding, B. Liu, S. Dang, Z. Lin, and J. Li.\ (2020) Privacy
Preservation in Location-Based Services: A Novel Metric and
Attack Model. \ {\it IEEE Transactions on Mobile Computing}, no. to appear.

[6] J. Konen, H. B. Mcmahan, D. Ramage, and P. Richtrik.\ (2016) \ Federated
optimization: Distributed machine learning for on-device intelligence.

[7] S. Shaham, M. Ding, B. Liu, S. Dang, Z. Lin, and J. Li.\ (2020) Privacy
Preserving Location Data Publishing: A Machine Learning Approach.
{\it IEEE Transactions on Knowledge and Data Engineering}, no. to appear.

[8] Q. Yang, Y. Liu, T. Chen, and Y. Tong.\ (2019) Federated machine
learning: Concept and applications. {\it Acm Transactions on Intelligent
Systems}, vol. 10, no. 2, pp. 12.1–12.19.

[9] D. Van Esch, E. Sarbar, T. Lucassen, J. O’Brien, T. Breiner,
M. Prasad, E. Crew, C. Nguyen, and F. Beaufays.\ (2019)  Writing across
the world’s languages: Deep internationalization for gboard, the
google keyboard. {\it arXiv}.

[10] C. Ma, J. Li, M. Ding, B. Liu, K. Wei, J. Weng, and
H. V. Poor.\ (2020) \ RDP-GAN: A R$\acute{e}$nyi-Differential Privacy based
Generative Adversarial Network. {\it arXiv}, [Online]. Available:
http://arxiv.org/abs/2007.02056

[11] K. Wei, J. Li, M. Ding, C. Ma, H. H. Yang, F. Farokhi, S. Jin, T. Q. S.
Quek, and H. Vincent Poor.\ (2020) \ Federated LearningWith Differential
Privacy: Algorithms and Performance Analysis. {\it IEEE Transactions
on Information Forensics and Security}, vol. 15, pp. 3454–3469.

[12] S. Nakamoto.\ (2009) \ Bitcoin: A peer-to-peer electronic cash system.
{\it Cryptography Mailing}, list at https://metzdowd.com.

[13] A. Reyna, C. Martłn, J. Chen, E. Soler, and M. Dłaz.\ (2018) \ On blockchain
and its integration with iot. challenges and opportunities. {\it Future
Generation Computer Systems}, vol. 88, no. NOV., pp. 173–190.

[14] H. Kim, J. Park, M. Bennis, and S. L. Kim.\ (2018) \ On-device federated
learning via blockchain and its latency analysis.

[15] I. Martinez, S. Francis, and A. S. Hafid.\ (2019)  Record and reward
federated learning contributions with blockchain. In {\it CyberC 2019
Workshop on Blockchain}.

[16] Y. Lu, X. Huang, Y. Dai, S. Maharjan, and Y. Zhang. \ (2019)  Blockchain
and federated learning for privacy-preserved data sharing in
industrial iot.\ {\it IEEE Transactions on Industrial Informatics}, vol. PP,
no. 99, pp. 1–1.

[17] S. Wang, T. Tuor, T. Salonidis, K. K. Leung, C. Makaya, T. He, and
K. Chan. \ (2018) \  Adaptive federated learning in resource constrained
edge computing systems.

[18] M. Chen, Z. Yang, W. Saad, C. Yin, H. V. Poor, and S. Cui. \ (2019) \ A joint
learning and communications framework for federated learning
over wireless networks.

[19] W. Gavin. \ (2014) \  Ethereum: a secure decentralised generalised transaction
ledger.

[20] A. Gervais, G. Karame, K. Wst, V. Glykantzis, H. Ritzdorf, and
S. Capkun. \ (2016) \ On the security and performance of proof of work
blockchains.

[21] C. Xu, K. Wang, P. Li, S. Guo, J. Luo, B. Ye, and M. Guo. \ (2019) \ Making
big data open in edges: A resource-efficient blockchain-based
approach. In {\it IEEE Transactions on Parallel and Distributed Systems},
vol. 30, no. 4, pp. 870–882.

[22]X. Deng, J. Li, L. Shi, Z. Wei, X. Zhou and J. Yuan, \ (2020) \ Wireless Powered Mobile Edge Computing: Dynamic Resource Allocation and Throughput Maximization. In {\it IEEE Transactions on Mobile Computing}, Early Access.

[23] H. Karimi, J. Nutini, and M. Schmidt. \ (2016) \ Linear convergence
of gradient and proximal-gradient methods under the polyaklojasiewicz
condition.

[24] Y. Lecun, L. Bottou, Y. Bengio, and P. Haffner. \ (1998) \ Gradient-based
learning applied to document recognition. \ {\it Proceedings of the IEEE},
vol. 86, no. 11, pp. 2278–2324.

\end{document}